%% file: main.tex
\documentclass[9pt,technote,compsoc]{IEEEtran}

\usepackage{url}
\usepackage{graphicx}  
\usepackage{amsfonts}
\usepackage{CJK}
\usepackage{amsmath}
\usepackage{bm}
\usepackage{mathrsfs}
\usepackage{float}
\usepackage{xcolor,colortbl}

\usepackage{subfigure}
\usepackage{booktabs,multirow}
\usepackage{bigstrut,bigdelim}
\usepackage{paralist}
\usepackage{bbm}

\usepackage{helvet} 
\usepackage{courier} 
\usepackage{makecell}
\usepackage{nicefrac}      

\usepackage{microtype} 
\usepackage{epsfig}
\usepackage{diagbox}

\usepackage{framed}
\usepackage{empheq}
\usepackage{array}
\usepackage{enumitem}
\usepackage{mdwlist}
\usepackage{xspace,mfirstuc,tabulary}

\usepackage{tcolorbox}

\newtcbox{\mybox}[1][red]
  {on line, arc = 0pt, outer arc = 0pt,
    colback = #1!10!white, colframe = #1!50!black,
    boxsep = 0pt, left = 1pt, right = 1pt, top = 2pt, bottom = 2pt,
    boxrule = 0pt, bottomrule = 1pt, toprule = 1pt}

\newcommand{\thickhline}{\noalign{\hrule height 1pt}}

\ifCLASSOPTIONcompsoc
  \usepackage[nocompress]{cite}
\else
  \usepackage{cite}
\fi

\ifCLASSINFOpdf
\else
\fi

\begin{document}

\title{Improving Matching Models with Hierarchical Contextualized Representations for Multi-turn Response Selection}

\author{Chongyang Tao, Wei~Wu, Can Xu, Yansong Feng, Rui Yan, and Dongyan~Zhao
\IEEEcompsocitemizethanks{

  \IEEEcompsocthanksitem Chongyang Tao, Yansong Feng, Rui Yan, and Dongyan Zhao are with the Institute of Computer Science \& Technology, Peking University, Beijing,
  China, 100871.  (e-mail: chongyangtao@pku.edu.cn; fengyansong@pku.edu.cn; ruiyan@pku.edu.cn; zhaody@pku.edu.cn)
  \IEEEcompsocthanksitem Wei Wu and Can Xu are with Microsoft Corporation, Beijing, China, 100871.  (e-mail: wuwei@microsoft.com; caxu@microsoft.com)
} 
}

\IEEEtitleabstractindextext{
\begin{abstract}
In this paper, we study context-response matching with pre-trained contextualized representations for multi-turn response selection in retrieval-based chatbots. Existing models, such as Cove and ELMo, are trained with limited context (often a single sentence or paragraph), and may not work well on multi-turn conversations, due to the hierarchical nature, informal language, and domain-specific words. To address the challenges, we propose pre-training hierarchical contextualized representations, including contextual word-level and sentence-level representations, by learning a dialogue generation model from large-scale conversations with a hierarchical encoder-decoder architecture. Then the two levels of representations are blended into the input and output layer of a matching model respectively. Experimental results on two benchmark conversation datasets indicate that the proposed hierarchical contextualized representations can bring significantly and consistently improvement to existing matching models for response selection. 
\end{abstract}

}

\maketitle

\IEEEdisplaynontitleabstractindextext
\IEEEpeerreviewmaketitle

\input{1-introduction}
\input{2-related}
\input{3-model}

\input{4-experiments}

\section{Conclusion and Future Work}
We propose pre-training a dialogue generation model from large-scale conversations with a hierarchical encoder-decoder architecture, and extracting both local and global contextualized word representations from the model to enhance matching models for multi-turn response selection. Experimental results on two benchmark datasets indicate that the proposed method can bring significant and consistent improvement to the performance of the existing matching models.

\ifCLASSOPTIONcaptionsoff
  \newpage
\fi

\bibliographystyle{IEEEtran}
\bibliography{main}

\end{document}

%% file: 1-introduction.tex
\section{Introduction} \label{sec:introduction}

Human-machine conversation is one of the fundamental problems in natural language processing (NLP). While previous research focuses on building task-oriented dialog systems~\cite{young2010hidden} that can fulfill specific tasks in vertical domains for people via conversations; more recent attention is drawn to developing non-task-oriented chatbots which can naturally and meaningfully converse with humans on open domain topics~\cite{shangL2015neural}. Existing approaches to building a chatbot include generation-based methods~\cite{shangL2015neural,sordoni2015neural,serban2015building,xing2017topic,serban2017hierarchical} which synthesize a response with natural language generation technologies, and retrieval-based methods~\cite{lowe2015ubuntu,wu2017sequential} which select a response from a pool of candidates.  In this work, we study multi-turn response selection for retrieval-based chatbots, because retrieval-based methods can return fluent and informative responses, and are the core of many real products such as the social-bot XiaoIce from Microsoft~\cite{shum2018eliza} and the E-commerce assistant AliMe Assist from Alibaba Group~\cite{li2017alime}. 

A key step to multi-turn response selection is measuring the matching degree between a conversational context consisting of a sequence of utterances and a response candidate with a matching model. Existing models, such as dual LSTM \cite{lowe2015ubuntu}, Multi-view \cite{zhou2016multi} and sequential matching network (SMN)~\cite{wu2017sequential} are defined in neural architectures. Although these models vary in structures, they are commonly built upon word embeddings which are pre-trained on large-scale unlabeled text with algorithms such as Word2Vec~\cite{mikolov2013distributed} and GloVe~\cite{pennington2014glove}. Indeed, the pre-trained word embeddings are crucial to a matching model, as they carry useful syntactic and semantic information learned from the unlabeled text to the matching task. On the other hand, words appear in specific contexts (e.g., sentences), and the same word could have different meanings in different contexts. The widely used embeddings, however, represent words in a context-independent way. As a result, contextual information of words in the unlabeled text is lost in the matching task.

In this work, we study how to leverage pre-trained contextualized representations to improve matching models for multi-turn response selection in retrieval-based chatbots. A baseline method is integrating the state-of-the-art contextualized word vectors such as CoVe~\cite{mccann2017learned} and ELMo~\cite{peters2018deep} into matching models.
Although both CoVe and ELMo have proven effective on various NLP tasks, 
they are never applied to conversation tasks. Therefore, it is not clear if CoVe and ELMo are as effective on the task of response selection as they are on other tasks such as machine reading comprehension~\cite{rajpurkar2016squad} and sentiment analysis~\cite{socher2013recursive}, etc. On the other hand, directly applying CoVe or ELMo to conversation modeling might be problematic, as there is a discrepancy between conversation data and the data used to train the two models. First, conversation data is often in a hierarchical structure, and thus there are both sentence-level contexts and session-level contexts for a word. A sentence-level context refers to an utterance that contains the word, and represents a kind of local context, and a session-level context means the entire conversation session (i.e., all utterances of the conversation history in question) that contains the word, and provides a global context for the word. CoVe and ELMo, however, only encode local contextual information of words when applied to a conversation task. Second, word use in conversation data is different from that in the WMT data which are used to train CoVe and ELMo. Words in conversations could be informal (e.g., ``srry'', which means ``sorry'' and is among the top $10$\% high-frequency words in Ubuntu dialogues) or domain-specific (e.g., ``fstab'', which means a system configuration file on Unix and Unix-like computer systems, and is among the top $1$\% high-frequency words in Ubuntu dialogues), and thus rarely appear in the WMT data. As a result, these words cannot be accurately represented by the two models. 
The discrepancy on data raises new challenges to leveraging contextualized word vectors in matching for response selection. 

To address the challenges, we propose pre-training contextualized representations with large-scale human-human conversations. Specifically, we employ a hierarchical encoder-decoder architecture, and learn a dialogue generation model with the conversations.  Local contextual information and global contextual information for a word are naturally encoded by the utterance-level recurrent neural network (RNN) and the context-level RNN of the encoder of the generation model respectively. Thus, we take the hidden states of the utterance-level RNN and the context-level RNN as contextualized word-level and sentence-level representations respectively, and name them ECMo (embeddings from a conversation model) representations. In matching, we integrate the word-level representation into the input layer of a matching model where words are initialized, and exploit the sentence-level representation in the output layer of the matching model where a matching score is calculated. By this means, we transfer the knowledge in large-scale unlabeled human-human conversations to the learning of the matching model.  

We integrate CoVe, ELMo, and ECMo into Multi-view and SMN, and conduct experiments on two benchmark datasets: the Ubuntu Dialogue Corpus~\cite{lowe2015ubuntu} and the Douban Conversation Corpus~\cite{wu2017sequential}. Experimental results indicate that the performance of matching models improves when they are combined with different contextualized representations (CoVe, ELMo (fine-tune) and ECMO). On the other hand, we observe that ECMo brings more significant and consistent improvement to matching models than CoVe and ELMo. On the Ubuntu data, the improvements to  Multi-view and SMN on R$_{10}$@1 are $4.3\%$ and $2.4\%$ respectively; and on the Douban data, the improvements to Multi-view and SMN on MAP are $2.0\%$ and $2.3\%$ respectively.

Our contributions are three-fold: 
\begin{itemize} 
    \item[1)] We test CoVe and ELMo on benchmark datasets of response selection;
    \item[2)] We propose a new approach to pre-training hierarchical contextualized representations that can well adapt to the task of response selection; 
    \item[3)] We verify the effectiveness of the proposed model on the benchmark datasets of response selection.
\end{itemize}

%% file: 2-related.tex
\section{Related Work}
Existing methods on building a chatbot are either generation-based or retrieval-based. The generation-based methods synthesize a response with the natural language generation technologies~\cite{serban2015building,shangL2015neural}. Different from generation-based methods, retrieval-based methods focus on designing a matching model of a human input and a response candidate for response selection. Early work along this line studies single-turn response selection where the human input is set as a single message~\cite{wang2013dataset,wang2015syntax}. Recently more attention is paid to context-response matching for multi-turn response selection. Representative methods include the dual LSTM model~\cite{lowe2015ubuntu}, the deep learning to respond architecture~\cite{rui2018learning}, the multi-view matching model~\cite{zhou2016multi}, the sequential matching network~\cite{wu2017sequential}, and the deep attention matching network~\cite{zhou2018multi}. In this work, we study the problem of multi-turn response selection for retrieval-based chatbots. Rather than designing a sophisticated matching structure, we are interested in how to leverage pre-trained contextualized word embeddings to generally improve the performance of the existing matching models. The contextualized embeddings are obtained from a dialogue generation model learned with large-scale human-human conversations apart from the training data of the matching models.

Pre-trained word vectors have become a standard component of most state-of-the-art models in NLP tasks. A common practice is to learn a single context-independent word representation from large-scale unlabeled data~\cite{mikolov2013distributed,pennington2014glove,bojanowski2017enriching}, and initialize the task-specific models with the representation. Recently, researchers begin to study pre-training context-dependent word representations for downstream tasks~\cite{melamud2016context2vec,peters2017semi,ramachandran2017unsupervised}. For example, McCann et al.~\cite{mccann2017learned} train an encoder-decoder model on large-scale machine translation datasets, and treat the hidden states of the encoder as contextualized representations of words. Peters et al.~\cite{peters2018deep} learn a multi-layer LSTM based language model on large-scale monolingual data, and use hidden states of different layers of the LSTM as contextualized word vectors. Very recently, Devlin et al.~\cite{devlin2018bert} propose a bigger and more powerful pre-trained model based on stacked self-attention. In this work, we study how to pre-train contextualized word representations for the task of multi-turn response selection which is never explored before. In addition to the application of CoVe and ELMo, we propose pre-training hierarchical contextualized word representations by learning a dialogue generation model from large-scale conversations with a hierarchical encoder-decoder. Different from the existing models, the dialogue generation model allows us to form two levels of contextualized word representations that encode both local and global contextual information for a word in conversations.   

%% file: 3-model.tex
\section{Background: Learning a Matching Model for Response Selection} 
Given a dataset $\mathcal {D} = \{(y_i,s_i,r_i)\}_{i=1}^N$ where $s_i=\{u_{i,1}, \ldots, u_{i,n_i}\}$ represents a conversational context with $\{u_{i,k}\}_{k=1}^{n_i}$ as utterances;  $r_i$ is a response candidate; and $y_i\in \{0,1\}$ denotes a label with $y_i=1$ indicating $r_i$ a proper response for $s_i$ and otherwise $y_i=0$, the goal of the task of response selection is to learn a matching model $g(\cdot,\cdot)$ from $\mathcal{D}$. For any context-response pair $(s,r)$, $g(s,r)$ gives a score that reflects the matching degree between $s$ and $r$, and thus allows one to rank a set of response candidates according to the scores for response selection.

For any  $u_{i,k}$ in $s_i$ and $r_i$ in $\mathcal{D}$, suppose that $u_{i,k}=(w_{i,k,1},\ldots,$ $w_{i,k, n_{i,k}})$ and $r_i=(v_{i,1}, \ldots, v_{i,n_i})$, where $w_{i,k,j}$ and $v_{i,j}$ denote the $j$-th words of $u_{i,k}$ and $r_i$ respectively, a common practice of learning of $g(\cdot,\cdot)$ is that $w_{i,k,j}$ and $v_{i,j}$ are first initialized with some pre-trained word vectors and then fed to an neural architecture. With the word vectors either fixed or optimized together with other parameters of the neural architecture, $g(\cdot,\cdot)$ is learnt by maximizing the following objective:
\begin{equation}
\label{oriobj}
\sum_{i=1}^{N} \big[ y_i \log(g(c_i,r_i))  + (1-y_i)\log(1-g(c_i,r_i))\big].
\end{equation} 

Existing work pre-trains word vectors with either Word2Vec (e.g., SMN \cite{wu2017sequential}) or GloVe (e.g., dual LSTM~\cite{lowe2015ubuntu} and multi-view \cite{zhou2016multi}) which loses contextual information in the representations of words. Therefore, inspired by the recent success of CoVe \cite{mccann2017learned} and ELMo \cite{peters2018deep} on downstream NLP tasks,  we consider incorporating contextualized word representations into the learning of $g(\cdot,\cdot)$. On the other hand, contexts of a word in conversations are usually in casual language, and sometimes with domain-specific knowledge (e.g., the Ubuntu Dialogue Corpus \cite{lowe2015ubuntu}). The contexts are naturally in a hierarchical structure where utterances and conversation sessions (i.e., sequences of utterances) that contain the word provide contextual information from a local and a global perspective respectively. The characteristics of conversation data motivate us to learn new contextualized representations of words that can well adapt to the task of response selection.

\section{ECMo: Embedding from a Conversation Model}
Heading for contextualized word representations that can well capture semantics and syntax of conversations, we propose learning a dialogue generation model from large-scale human-human conversations. We first present the architecture of the generation model, and then elaborate how to extract two levels of contextualized representations from the model and exploit them in matching. 

\begin{figure}[t]	
	\begin{center}
		\includegraphics[width=0.4\textwidth]{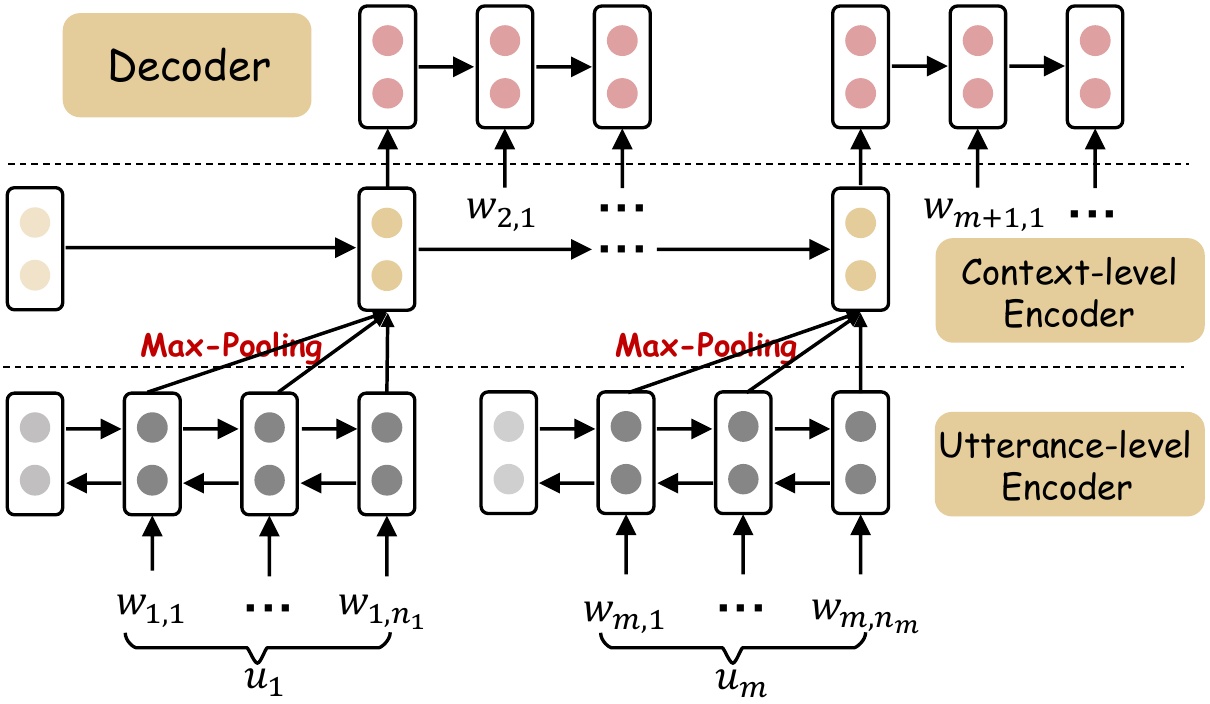}	
	\end{center}
	\caption{The architecture of HED.}
	\label{fig:workflow}
\end{figure}   

\subsection{Hierarchical Encoder-Decoder Model}
In order to represent contextual information in both utterances and the entire session of a conversation, we learn a hierarchical encoder-decoder (HED) model for multi-turn dialogue generation. Figure~\ref{fig:workflow} gives the architecture of HED. HED consists of a two-level encoder and a decoder. The first layer of the encoder is an utterance-level encoder where HED reads utterances in conversation history one by one, and represents the word sequence of each utterance as a sequence of hidden vectors by a bidirectional RNN with Gated Recurrent Units (biGRUs)~\cite{cho2014properties}. The hidden vectors of each utterance are then processed by a max pooling operation and transformed to an utterance vector. The second layer of the encoder is a context-level encoder which employs another GRU to transform the sequence of utterance vectors to hidden vectors. The final hidden vector of the context-level encoder is fed to the decoder to generate the next turn. 

Note that HED is in a similar structure with  the model proposed by~\cite{serban2015building}. The difference is that in HED, we represent each utterance with a biGRU instead of a GRU in order to encode both forward and backward contextual information in an utterance, and employ a max pooling operation to generate an utterance vector instead of only using the last hidden state, as suggested by~\cite{conneau2017supervised}. We select HED among many dialogue generation models because the two-layer encoder naturally encodes local and global contextual information in conversations, and the model balances efficacy and efficiency (to facilitate learning from large-scale conversations) compared to more complicated architectures such as VHRED~\cite{serban2017hierarchical}.

\paragraph{\textbf{Utterance-Level Encoder:}} given a sequence of utterances $s=\{u_1, \ldots, u_n\}$, we employ a biGRU to encode each $u_i$ as hidden vectors $\{\mathbf{h}_{i,k}\}_{k=1}^{T_i}$. Formally, suppose that $u_i=\{ w_{i,k}\}_{k=1}^{T_i}$, then $\forall k\in\{1,\ldots,T_i\}$, $\mathbf{h}_{i,k}$ is given by
\begin{equation} \label{eq: encoder function}
\mathbf{h}_{i,k}=[\overrightarrow{\mathbf{h}}_{i,k}; \overleftarrow{\mathbf{h}}_{i,k}],
\end{equation}
where $\overrightarrow{\mathbf{h}}_{i,k}$ is the $k$-th hidden state of a forward GRU \cite{cho2014properties} and $\overleftarrow{\mathbf{h}}_{i,k}$ is the $k$-th hidden state of a backward GRU. 

The output of the utterance-level encoder is a sequence of utterance vectors $\{\mathbf{v}^u_i\}_{i=1}^n$ where $\mathbf{v}^u_i$ is the representation of $u_i$ with the $j$-th element as
\begin{equation}
\mathbf{v}^u_i(j)  = \max(\mathbf{h}_{i,1}(j), \ldots, \mathbf{h}_{i,T_i}(j)),
\end{equation}
where $\mathbf{h}_{i,1}(j)$ and $\mathbf{h}_{i,T_i}(j)$ are the $j$-th elements of $\mathbf{h}_{i,1}$ and $\mathbf{h}_{i,T_i}$ respectively.

\paragraph{\textbf{Context-Level encoder:}} the context-level encoder takes the output of the utterance-level encoder as input, and represents the entire conversation session $s$ as a sequence of hidden vectors $\{\mathbf{h}^s_i\}_{i=1}^n$, where $\mathbf{h}^s_i$ is calculated by
\begin{equation} \label{contex encoder}
\mathbf{h}^s_i = \text{GRU}_\text{s}(\mathbf{h}^s_{i-1}, \mathbf{v}^u_i),
\end{equation}

\paragraph{\textbf{Decoder:}} the decoder of HED is an RNN language model~\cite{mikolov2010recurrent} which predicts the next utterance $u_{n+1}$ word by word conditioned on $\mathbf{h}^s_n$. Suppose that $u_{n+1}=\{ w_{n+1,k}\}_{k=1}^{T_{n+1}}$, then the generation probability of $p(u_{n+1}|u_1,\ldots, u_n) $ is defined as
\begin{equation} \label{eq:probability} 
\begin{aligned}
p(w_{n+1,1}|\mathbf{h}^s_n) \cdot \prod_{t=2}^{T_{n+1}} p(w_{n+1,t} | \mathbf{h}^s_n, \{w_{n+1,k}\}_{k=1}^{t-1}), 
\end{aligned}
\end{equation}
where $ p(w_{n+1,t} | \mathbf{h}^s_n, \{w_{n+1,k}\}_{k=1}^{t-1})$ is given by
\begin{equation}
\label{eq: decoder function}
\mathbbm{I}_{w_{n+1,t}} \cdot \text{softmax}(\mathbf{h}^d_t,\mathbf{e}_{n+1,t-1}).
\end{equation}
$\mathbf{h}^d_t$ is the hidden state of the decoder at step $t$ which is defined as
\begin{equation}
\mathbf{h}^d_t=\text{GRU}_\text{d}(\mathbf{h}^d_{t-1}, \mathbf{e}_{n+1,t-1}), 
\end{equation}
where $\mathbf{e}_{n+1,t-1}$ is the embedding of $w_{n+1,t-1}$, $\mathbbm{I}_{w_{n+1,t}}$ is a one-hot vector of $w_{n+1,t}$, and $\text{softmax}(\cdot,\cdot)$ is a $V$-dimensional vector ($V$ is the vocabulary size) of which each element is the generation probability of a word. We initialize the recurrent state of the \text{GRU}$_\text{d}$ with a nonlinear transformation of $\mathbf{h}^s_n$.

\paragraph{\textbf{Learning objective:}} we estimate the parameters of HED by maximizing the likelihood of a dataset $\mathcal{D}'=\{s_i\}_{i=1}^{N'}$ where $s_i$ is a conversation session. The source of $\mathcal{D}'$ could be different from that of $\mathcal{D}$, and $N'$ could be much larger than $N$, as will be seen in our experiments later. Thus, we can transfer the knowledge in large scale unlabeled conversations to the learning of a matching model. Suppose that $s_i=\left(u_{i,1},\ldots, u_{i,n_i}\right)$, then the learning of HED can be formulated as maximizing the following objective: 
\begin{equation}
\sum_{i=1}^{N'} \sum_{j=2}^{n_i} \log \big( p(u_{i,j} | u_{i,1}, \ldots, u_{i,j-1}) \big).
\end{equation}

\subsection{ECMo}
ECMo are representations defined by the hidden states of the two-level encoder of HED. Given a word $w_{i,k}$ in an utterance $u_i$ in a conversation session $s$, the word-level contextualized representation of $w_{i,k}$ is defined by
\begin{equation} \label{sentence_level}
\text{ECMo}_\text{local}(w_{i,k})= \mathbf{h}_{i,k},
\end{equation} 
where $\mathbf{h}_{i,k}$ is given by Equation (\ref{eq: encoder function}). The sentence-level contextualized representation of $w_{i,k}$ is defined by
\begin{equation} \label{session_level}
\text{ECMo}_\text{global}(w_{i,k})=\mathbf{h}^s_i,
\end{equation} 
where $\mathbf{h}^s_i$ is given by Equation (\ref{contex encoder}).

\subsection{Using ECMo for Matching Models}
Given a pre-trained HED and a matching model, we incorporate ECMo representations into both the input layer and the output layer of the matching model by running the encoder of the HED on a conversational context (i.e., a conversation session) and a response candidate (treated as a special conversation session with only one utterance) simultaneously. Existing matching models share a common architecture at the input layers where words are initialized with pre-trained vectors, and act in a common manner at the output layers where a context-response pair is transformed to a score, which allows us to add ECMo in a unified way. 

Formally, suppose that the input of a matching model $g(\cdot,\cdot)$ is a conversational context $s= \{ u_i\}_{i=1}^{n}$ with $u_i$ the $i$-th utterance and a response candidate $r$, let $u_i= \{ w_{i,k} \}_{k=1}^{n_i}$ and $r=\{ v_k\}_{k=1}^{n_r}$, where word $w_{i,k}$ and word $v_k$ are initialized by pre-trained context-independent representations $\mathbf{e}^u_{i,k}$ and $\mathbf{e}^r_k$ respectively, we then form a new representation for $w_{i,k}$ as
\begin{equation}
\tilde{\mathbf{e}}^u_{i,k}=[\mathbf{e}^u_{i,k}; \text{ECMo}_\text{local}(w_{i,k})], 
\end{equation}
Similarly, we form a new representation for $\mathbf{v}_k$ as
\begin{equation}
\tilde{\mathbf{e}}^r_k=[\mathbf{e}^r_k; \text{ECMo}_\text{local}(\mathbf{v}_k)].
\end{equation}
We then initialize the embedding of $w_{i,k}$ and ${v}_k$ at the input layer of $g(\cdot,\cdot)$ with $\tilde{\mathbf{e}}^u_{i,k}$ and $\tilde{\mathbf{e}}^r_k$ respectively. At the output layer of $g(\cdot, \cdot)$, in addition to $g(s,r)$, we define a new matching score based on the sentence-level contextualized representations, which is defined as
\begin{equation}
g'(s,r)=\sigma(\tilde{\mathbf{e}}^s \cdot \mathbf{W} \cdot \tilde{\mathbf{e}}^{r}+ \mathbf{b}),
\end{equation}
where $\tilde{\mathbf{e}}^s= \text{ECMo}_\text{global}(w_{n,k})$, $\tilde{\mathbf{e}}^{r}= \text{ECMo}_\text{global}(v_{n_r})$, and $\mathbf{W}$ and $\mathbf{b}$ are parameters, . We then re-define the matching model $g(s,r)$ as
\begin{equation}
\tilde{g}(s,r)=g(s,r)+g'(s,r),
\end{equation}

In learning of the matching model, one can either freeze the parameters of the pre-trained HED or continue to optimize those parameters with the cross entropy loss given by Equation (\ref{oriobj}). We empirically compare the two strategies in our experiments.

%% file: 4-experiments.tex
\section{Experiments}
We test CoVe, ELMo, and ECMo on two benchmark datasets for multi-turn response selection.
\subsection{Experiment Setup}
\paragraph{Ubuntu Dialogue Corpus:} the Ubuntu Dialogue Corpus~\cite{lowe2015ubuntu} is an English dataset collected from chat logs of the Ubuntu Forum.  We use the version provided by \cite{xu2016incorporating} (i.e., Ubuntu Dialogue Corpus v1). There are $1$ million context-response pairs for training, $0.5$ million pairs for validation, and $0.5$ million pairs for the test. In the data, responses from humans are treated as positive responses, and negative responses are randomly sampled. In the training set, the ratio of positive responses and negative responses is 1:1. In the validation set and the test set, the ratios are 1:9. Following~\cite{lowe2015ubuntu}, we employ $R_n@k$s as evaluation metrics.

\paragraph{Douban Conversation Corpus:} the Douban Conversation Corpus~\cite{wu2017sequential} is a multi-turn Chinese conversation dataset crawled from Douban group\footnote{\url{https://www.douban.com/group}}. The dataset consists of $1$ million context-response pairs for training, $50$ thousand pairs for validation, and $6,670$ pairs for the test. In the training set and the validation set, the last turn of each conversation is regarded as a positive response and the negative responses are randomly sampled. The ratio of positive responses and negative responses is 1:1 in training and validation. In the test set,  each context has $10$ response candidates retrieved from an index whose appropriateness regarding to the context is judged by human labelers. Following \cite{wu2017sequential}, we also employ $R_n@k$s, mean average precision (MAP), mean reciprocal rank (MRR) and precision at position 1~(P@1) as evaluation metrics.

\begin{table*}[h!]
    \centering
    \caption{Evaluation results on the two datasets. Numbers in bold mean that improvement to the original models brought by ECMo is statistically significant (t-test, p-value $<0.01$ ). Numbers marked with $^*$ mean that improvement to ELMo (fine-tune) and CoVe is statistically significant  (t-test, p-value $<0.01$ ). We do not include the results of CoVe and ELMo enhanced models on the Douban data because the two models are not available for Chinese data.}   
    \resizebox{\textwidth}{!}{
        \begin{tabular}{l|c|c|c|c||c|c|c|c|c|c}
            \thickhline \hline &   \multicolumn{4}{c||}{\textbf{Ubuntu Corpus}} &  \multicolumn{6}{c}{\textbf{Douban Corpus}} \\ \cline{2-11}
            &  $R_2@1$      &  $R_{10}@1$ &  $R_{10}@2$ &  $R_{10}@5$ & MAP & MRR & P@1 & $R_{10}@1$ & $R_{10}@2$ & $R_{10}@5$\\ \hline
            Multi-view & 0.916 & 0.690 & 0.831 & 0.959 & 0.502 & 0.547 & 0.352 & 0.205 & 0.353 & 0.728  \\ 
            Multi-view + CoVe & 0.919 & 0.699 & 0.837 & 0.960 & - & - & - & - & - & - \\ 
            Multi-view + ELMo & 0.909 & 0.668 & 0.813 & 0.951 & - & - & - & - & - & -\\ 
            Multi-view + ELMo (fine-tune) & 0.924 & 0.705 & 0.847 & 0.964 & - & - & - & - & - & -\\
            Multi-view + ECMo & \textbf{0.930} & \textbf{0.733}$^*$ & \textbf{0.858}$^*$ & \textbf{0.967} & \textbf{0.522} & \textbf{0.572} & \textbf{0.378} & \textbf{0.224} & \textbf{0.391} & \textbf{0.761} \\ \hline
                        SMN & 0.925 & 0.732 & 0.852 & 0.961 & 0.526 & 0.571 & 0.393 & 0.236 & 0.387 & 0.729 \\ 
            SMN + CoVe & 0.930 & 0.738 & 0.856 & 0.963 & - & - & - & - & - & - \\
            SMN + ELMo & 0.926 & 0.739 & 0.855 & 0.961 & - & - & - & - & - & -\\ 
            SMN + ELMo (fine-tune) & 0.930 & 0.745 & 0.859 & 0.963 & - & - & - & - & - & -\\           
            SMN + ECMo& \textbf{0.934} & \textbf{0.756}$^*$ & \textbf{0.867}$^*$ & \textbf{0.966} & \textbf{0.549} & \textbf{0.593} & \textbf{0.409} & 0.247 & \textbf{0.416} & \textbf{0.774}\\ \hline
            \thickhline
        \end{tabular}
    }
    \label{tab:main} 
\end{table*}

\subsection{HED Pre-training}
The HED models for both datasets are trained using Adam~\cite{kingma2014adam} with a mini-batch $40$. The learning rate is set as $1\text{e}^{-3}$. The size of the hidden vectors of the utterance-level RNN, the context-level RNN, and the decoder RNN are $300$. Since the utterance-level RNN is bidirectional, the dimension of ECMo$_{\text{local}}$ vectors is $600$, which is the same as CoVe. We set the maximum length of a session as $10$ and the maximum length of an utterance as $50$.  

For the Ubuntu data, we crawl $10$ million multi-turn conversations from Twitter, covering 2-month period from June 2016 to July 2016. As pre-processing, we remove URLs, emotions, and usernames, and transform each word to lower case. On average, each conversation has $9.2$ turns.  
The HED model is first trained on the Twitter data, and then is fine-tuned on the training set of the Ubuntu data. By this means, we encode both the semantics in the casual conversations of Twitter and the domain knowledge in the Ubuntu dialogues. In training, we initialize word embeddings with $300$-dimensional GloVe vectors~\cite{pennington2014glove}.  The vocabulary is constructed by merging the vocabulary of the Ubuntu data (the size is $60$k) and the vocabulary of the Twitter data (the size is $60$k). The size of the final vocabulary is $99,394$.
Words that are out of the vocabulary are randomly initialized according to a zero-mean normal distribution. After the first stage of training, the HED model achieves a perplexity of $70.3$ on a small validation set of the Twitter data ($20$k). The perplexity of the final HED model on the validation set of the Ubuntu data is $59.4$. 

For the Douban data, we train the HED model on a public dataset\footnote{Avalible at \url{http://tcci.ccf.org.cn/conference/2018/dldoc/trainingdata05.zip}}. The dataset contains $5$ million conversations crawled from Sina Weibo\footnote{\url{www.weibo.com}}. On average, each conversation has $4.1$ turns. Since conversations from Weibo cover a wide range of topics which are similar to those in the Douban Conversation Corpus, we only train the HED model on the Weibo data. Word embeddings are initialized by running Word2Vec \cite{mikolov2013distributed} on the Weibo data. The final model achieves a perplexity of $123.7$ on the validation set of the Douban data.

\subsection{Matching Models}
The following matching models are selected to test the effect of pre-trained contextualized word representations:

\textbf{Multi-view}: the model proposed by~\cite{zhou2016multi} in which the response vector interacts with a highly abstract context vector obtained from both word view and utterance view. We select the model as it is representative architecture for context-response matching and is better than dual LSTM~\cite{lowe2015ubuntu}.

\textbf{Sequential matching network (SMN)}: the model proposed by \cite{wu2017sequential} in which each utterance in a context interacts with a response word-by-word at the beginning, and the interaction is transformed to a matching vector by 2D CNNs. The matching vectors are finally aggregated by an RNN as a matching score. We select the model as it is a representative in the framework of representation-matching-aggregation for context-response matching.


\begin{table*}[t!] 
\centering
\caption{Evaluation results of model ablation. }
\resizebox{\textwidth}{!}{
    \begin{tabular}{l|c|c|c|c||c|c|c|c|c|c}
        \thickhline &   \multicolumn{4}{c||}{\textbf{Ubuntu Corpus}} & \multicolumn{6}{c}{\textbf{Douban Corpus}}  \\ \cline{2-11}
        &  R$_2$@1      &  R$_{10}$@1 &  R$_{10}$@2&  R$_{10}$@5 &MAP&MRR&P@1  &  $R_{10}@1$ & $R_{10}@2$ & $R_{10}@5$  \\ \hline
        Multi-view + ECMo$_L$ & 0.921 & 0.698 & 0.837 & 0.961 & 0.510 & 0.554 & 0.358 & 0.210 & 0.380 & 0.731  \\ 
        Multi-view + ECMo$_G$ & 0.931 & 0.729 & 0.858 & 0.967 & 0.515 & 0.562 & 0.373 & 0.216 & 0.384 & 0.755\\ 
        Multi-view + ECMo$_L$ (twitter) & 0.916 & 0.686 & 0.829 & 0.959 & - & - & - &- & - & -\\
        Multi-view + ECMo$_G$ (twitter) & 0.920 & 0.704 & 0.839 & 0.960 & - & - & - &- & - & -\\ 
        Multi-view + ECMo (twitter) & 0.918 & 0.699 & 0.838 & 0.960 & - & - & - &- & - & -\\ 
        Multi-view + ECMo (Ubuntu-only) & 0.915 & 0.686 & 0.826 & 0.956 & - & - & - &- & - & -\\ \hline
        Multi-view + ECMo & 0.930 & {0.733} & {0.858} & 0.967 & {0.522} & {0.572} & {0.378} & {0.224} & {0.391} & {0.761} \\ \thickhline        
        SMN + ECMo$_L$ & 0.932 & 0.747 & 0.862 & 0.966 & 0.540 & 0.584 & 0.399 & 0.240 & 0.401 & 0.759  \\ 
        SMN + ECMo$_G$ & 0.932 & 0.750 & 0.864 & 0.964 & 0.537 & 0.582 & 0.396 & 0.233 & 0.406 & 0.753\\ 
        SMN + ECMo$_L$ (twitter) & 0.929 & 0.743 & 0.859 & 0.962 & - & - & - &- & - & -\\
        SMN + ECMo$_G$ (twitter) & 0.917 & 0.718 & 0.841 & 0.951 & - & - & - &- & - & -\\ 
        SMN + ECMo (twitter) & 0.918 & 0.719 & 0.844 & 0.953 & - & - & - &- & - & -\\ 
        SMN + ECMo (Ubuntu-only) & 0.917 & 0.717 & 0.839 & 0.948 & - & - & - &- & - & -\\ \hline
        SMN + ECMo & 0.933 & 0.755 & 0.866 & 0.975 & 0.549 & 0.593 & 0.409 & 0.247 & 0.416 & 0.774 \\ 
        \thickhline
    \end{tabular}
}
\label{fig:ablation}    
\end{table*}

For CoVe and ELMo, we use the models published at \url{https://github.com/salesforce/cove} and \url{https://github.com/allenai/allennlp/blob/master/tutorials/how_to/elmo.md} respectively, and follow the way in the existing papers to integrate the contextualized vectors into the wording embedding layer of the matching models. The dimensions of the vectors from CoVe and ELMo are $600$ and $1024$ respectively. The combination weights of ELMo are optimized together with the matching models, as suggested in \cite{peters2018deep}. All parameters of the pre-trained models are fixed during the learning of the matching models. Since HED is fine-tuned on the Ubuntu data, to make a fair comparison, we also fine-tune ELMo on the Ubuntu data\footnote{We do not fine-tune CoVe because CoVe needs paired data and thus is difficult to fine-tune with the conversation data.}.

To integrate the pre-trained contextualized vectors, we implement Multi-view and SMN in the same setting as in \cite{zhou2016multi} and \cite{wu2017sequential} respectively with PyTorch 0.3.1. Note that as indicated in \cite{zhou2016multi,wu2017sequential}, the context-independent word vectors (i.e., GloVe in Multi-view and Word2Vec in SMN) are learned or fine-tuned with the training sets of the Ubuntu and the Douban data, and the dimension of the vectors is $200$.

\subsection{Evaluation Results}
Table~\ref{tab:main} reports the evaluation results on the two datasets. We can observe that the performance of Multi-view and SMN improves on almost all metrics after they are combined with CoVe, ELMo (fine-tune), and ECMo, indicating that contextualized vectors are useful to the multi-turn response selection task. Notably, ECMo brings more significant and more consistent improvement to matching models on both datasets, which verifies the effectiveness of the proposed method. With ECMo, a simple Multi-view even performs better than SMN on the Ubuntu data, although SMN is in a more complicated structure, and thus is capable of capturing more semantic information in conversational contexts. 

Without fine-tuning, the performance of Multi-view drops with ELMo while the performance of SMN slightly increases. The reason might be that Multi-view is in a relatively simple structure and ELMo (w/o finetune) introduces high-dimensional discrepant word representations and also brings too many parameters.

\begin{table*}[h!] 
\centering
\caption{The results of the models with fixed or continue-trained ECMo representations.}
\resizebox{\textwidth}{!}{
    \begin{tabular}{l|c|c|c|c||c|c|c|c|c|c}
        \thickhline &   \multicolumn{4}{c||}{\textbf{Ubuntu Corpus}} &  \multicolumn{6}{c}{\textbf{Douban Corpus}}  \\ \cline{2-11}
        &  R$_2$@1      &  R$_{10}$@1 &  R$_{10}$@2&  R$_{10}$@5 &MAP&MRR&P@1 &  $R_{10}@1$ & $R_{10}@2$ & $R_{10}@5$ \\\hline
            Multi-view + ECMo & 0.930 & {0.733} & {0.858} & 0.967 & {0.522} & {0.572} & {0.378} & {0.224} & {0.391} & {0.761} \\
            Multi-view + ECMo (continue-train)& 0.929 & 0.723 & 0.855 & 0.966  & 0.516 & 0.558 &0.367 & 0.222 & 0.372 & 0.739  \\ \hline
            SMN + ECMo & 0.933 & 0.755 & 0.866 & 0.975 & 0.549 & 0.593 & 0.409 & 0.247 & 0.416 & 0.774 \\
            SMN + ECMo (continue-train) &0.935 & 0.749 & 0.866 & 0.968 & 0.544 & 0.587 & 0.406 & 0.246 & 0.414 & 0.765 \\ 
        \thickhline
    \end{tabular}
}
\label{fig:finetune}
\end{table*}

\subsection{Analysis}
\textbf{Model ablation:} we investigate how different configurations of ECMo affect the performance of matching through an ablation study. We first check how the word-level and the sentence-level contextualized representations individually contribute to the improvement on matching performance by leaving only one type of representations, and denote the models as model+ECMo$_L$ and model+ ECMo$_G$ respectively. Then, we examine if fine-tuning the HED model on the Ubuntu training data after it is trained on the Twitter data matters a lot by integrating the ECMo representations from the HED model only trained on the Twitter data into Multi-view and SMN. The models are denoted as model+ECMo$_L$ (twitter), model+ECMo$_G$ (twitter), and model + ECMo (twitter) respectively. Finally, we are also curious about if HED is already good enough when it is only pre-trained with the Ubuntu training data. Models that are combined with such ECMo representations are denoted as model+ECMo (Ubuntu-only).

Table~\ref{fig:ablation} reports the results. First, we find that both ECMo$_L$ and ECMo$_G$ are useful to matching. ECMo$_G$ generally brings more improvement to matching than ECMo$_L$ for Multi-view while the results reverse for SMN. This phenomenon is mainly due to the different interaction operation of the two models. In Multi-view, the response and context are interacted with corresponding global utterance representations, while in SMN, the interaction is performed word-by-word.
Second, fine-tuning is necessary as the performance of models drops without it. 
Since the Ubuntu dialogues have some domain-specific knowledge, a HED model only trained on Twitter cannot fully capture the contextual information in the Ubuntu data, resulting in inaccurate contextualized word representations. We can see that ECMo$_G$ and ECMo$_L$ shows different robustness to the noise for Multi-view and SMN due to the different interaction operation. 
Finally, although fine-tuning is important, one cannot only train a HED model on the domain-specific data, because the size of the data is small and a lot of contextual information in common conversation language will be lost. That explains why model+ECMo (Ubuntu-only) is inferior to all other variants.

\textbf{Does further optimization under the matching loss help?} So far, we fix the pre-trained ECMo representations in matching. Then it is interesting to know if we can obtain more improvement when we continue to train the parameters of HED under the cross entropy objective (i.e., Objective (\ref{oriobj})) of matching.  Table~\ref{fig:finetune} compares the two settings where models whose ECMo is optimized under the matching objective are denoted as model+ECMo (continue-train). We find that ``continue-train'' makes the performance of Multi-view and SMN drop on both datasets. This phenomenon may stem from the fact that the matching model is prone to over-fitting with the additional parameters coming from HED. On the other hand, fine-tuning with cross-entropy loss may damnify the original contextualized representations.